\documentclass[10pt,twocolumn,letterpaper]{article}

\usepackage{cvpr}
\usepackage{times}
\usepackage{epsfig}
\usepackage{graphicx}
\usepackage{amsmath}
\usepackage{amssymb}

\usepackage{multirow}
\usepackage[usestackEOL]{stackengine}
\usepackage{subfigure}

\usepackage{caption}
\newcommand{\squishlistbegin}{
 \begin{list}{$\bullet$}
  { \setlength{\itemsep}{0pt}
     \setlength{\parsep}{2pt}
     \setlength{\topsep}{2pt}
     \setlength{\partopsep}{0pt}
     \setlength{\leftmargin}{1.5em}
     \setlength{\labelwidth}{1em}
     \setlength{\labelsep}{0.5em} 
  } 
}
\newcommand{\squishlistend}{\end{list}}
\usepackage{array}
\newcolumntype{L}[1]{>{\raggedright\let\newline\\\arraybackslash\hspace{0pt}}m{#1}}
\newcolumntype{C}[1]{>{\centering\let\newline\\\arraybackslash\hspace{0pt}}m{#1}}
\newcolumntype{R}[1]{>{\raggedleft\let\newline\\\arraybackslash\hspace{0pt}}m{#1}}

\usepackage[pagebackref=true,breaklinks=true,letterpaper=true,colorlinks,bookmarks=false]{hyperref}

\cvprfinalcopy 

\ifcvprfinal\fi
\begin{document}


\title{PackNet: Adding Multiple Tasks to a Single Network by Iterative Pruning}

\author{Arun Mallya and Svetlana Lazebnik\\
University of Illinois at Urbana-Champaign\\
{\tt\small \{amallya2,slazebni\}@illinois.edu}
}


\maketitle


\begin{abstract}
This paper presents a method for adding multiple tasks to a single deep neural network while avoiding catastrophic forgetting. 
Inspired by network pruning techniques, we exploit redundancies in large deep networks to free up parameters that can then be employed to learn new tasks. 
By performing iterative pruning and network re-training, we are able to sequentially ``pack'' multiple tasks into a single network while ensuring minimal drop in performance and minimal storage overhead.
Unlike prior work that uses proxy losses to maintain accuracy on older tasks, we always optimize for the task at hand.
We perform extensive experiments on a variety of  network architectures and large-scale datasets, and observe much better robustness against catastrophic forgetting than prior work. 
In particular, we are able to add three fine-grained classification tasks to a single ImageNet-trained VGG-16 network and achieve accuracies close to those of separately trained networks for each task. 
Code available at \url{https://github.com/arunmallya/packnet}
\end{abstract}

\section{Introduction}
\label{sec:Introduction}

Lifelong or continual learning~\cite{aljundi2016expert,kirkpatrick2017overcoming,rannen2017encoder} is a key requirement for general artificially intelligent agents. Under this setting, the agent is required to acquire expertise on new tasks while maintaining its performance on previously learned tasks, ideally without the need to store large specialized models for each individual task. In the case of deep neural networks, the most common way of learning a new task is to fine-tune the network. However, as features relevant to the new task are learned through modification of the network weights, weights important for prior tasks might be altered, leading to deterioration in performance referred to as \lq\lq\emph{catastrophic forgetting}\rq\rq~\cite{french1999catastrophic}. 
Without access to older training data due to the lack of storage space, data rights, or deployed nature of the agent, which are all very realistic constraints, na\"ive fine-tuning is not a viable option for continual learning.

\begin{figure*}[t!]
  \centering
  \includegraphics[trim={0cm 0.5cm 0cm 0cm},width=\textwidth]{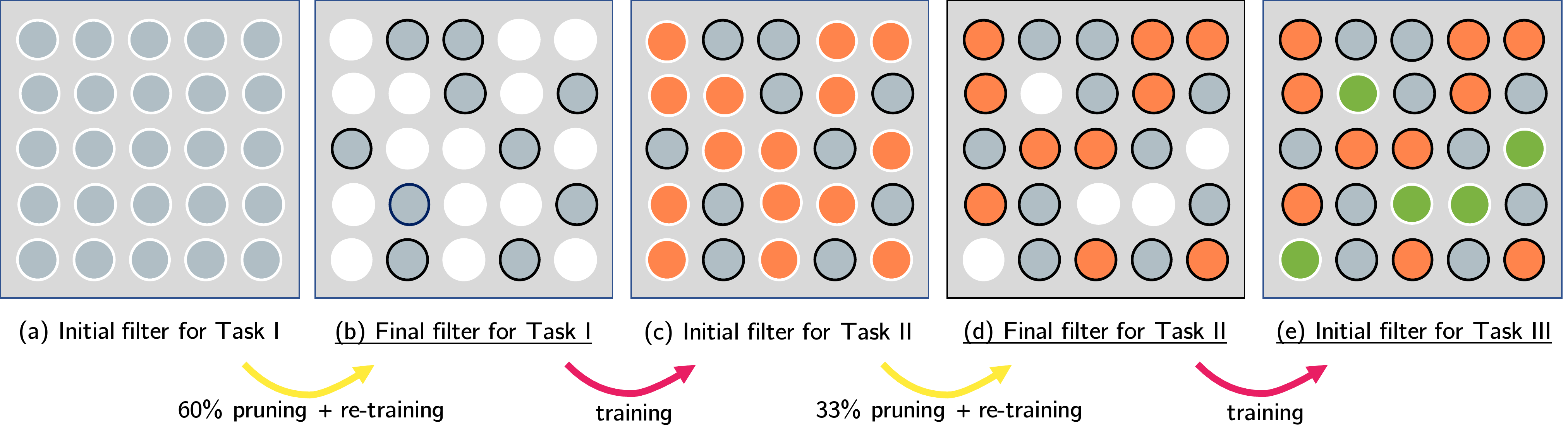}
  \caption{Illustration of the evolution of a 5$\times$5 filter with steps of training. Initial training of the network for Task I 
  learns a dense filter as illustrated in (a). After pruning by 60\% (15/25) and re-training, we obtain a sparse 
  filter for Task I, as depicted in (b), where white circles denote 0 valued weights. Weights retained 
  for Task I are kept fixed for the remainder of the method, and are not eligible for further pruning. We allow the pruned weights to be updated for Task II, leading to filter (c), which shares weights learned for Task I. Another round of pruning by 33\% (5/15) and re-training leads to filter (d), which is the filter used for evaluating on task II (Note that weights for Task I, in gray, are not considered for pruning). Hereafter, weights for Task II, depicted in orange, are kept fixed.
  This process is completed until desired, or we run out of pruned weights, as shown in filter (e). The final filter (e) for task III shares weights learned for tasks I and II. At test time, appropriate masks are applied depending on the selected task so as to replicate filters learned for the respective tasks.}
  \label{fig:filters_illustration}
\end{figure*}

Current approaches to overcoming catastrophic forgetting, such as Learning without Forgetting (LwF) ~\cite{li2016learning} and Elastic Weight Consolidation (EWC)~\cite{kirkpatrick2017overcoming}, have tried to preserve knowledge important to prior tasks through the use of proxy losses. The former tries to preserve activations of the initial network while training on new data, while the latter penalizes the modification of parameters deemed to be important to prior tasks. 
Distinct from such prior work, we draw inspiration from approaches in network compression that have shown impressive results for reducing network size and computational footprint by eliminating redundant parameters~\cite{han2015learning,li2016pruning,liu2017learning,molchanov2016pruning}.
We propose an approach that uses weight-based pruning techniques~\cite{han2016dsd,han2015learning} to free up redundant parameters across all layers of a deep network after it has been trained for a task, with minimal loss in accuracy. Keeping the surviving parameters fixed, the freed up parameters are modified for learning a new task. This process is performed repeatedly for adding multiple tasks, as illustrated in Figure~\ref{fig:filters_illustration}. By using the task-specific parameter masks generated by pruning, our models are able to maintain the same level of accuracy even after the addition of multiple tasks, and incur a very low storage overhead per each new task.

Our experiments demonstrate the efficacy of our method on several tasks for which high-level feature transfer does not perform very well, indicating the need to modify parameters of the network at all layers. In particular, we take a single ImageNet-trained VGG-16 network~\cite{simonyan14VGG} and add to it three fine-grained classification tasks 
 -- CUBS birds~\cite{WahCUB_200_2011}, Stanford Cars~\cite{krause20133d}, and Oxford Flowers~\cite{Nilsback08} -- while achieving accuracies very close to those of separately trained networks for each individual task. This significantly outperforms prior work in terms of robustness to catastrophic forgetting, as well as the number and complexity of added tasks. We also show that our method is superior to joint training when adding the large-scale Places365~\cite{zhou2017places} dataset to an ImageNet-trained network, and obtain competitive performance on a broad range of architectures, including VGG-16 with batch normalization~\cite{ioffe2015batch}, ResNets~\cite{he2016deep}, and DenseNets~\cite{huang2017densely}.

\section{Related Work}
\label{sec:related_work}

A few prior works and their variants, such as Learning without Forgetting (LwF)~\cite{li2016learning,rannen2017encoder,shmelkov2017incremental} and Elastic Weight Consolidation (EWC)~\cite{kirkpatrick2017overcoming,lee2017overcoming}, are aimed at training a network for multiple tasks sequentially. When adding a new task, LwF preserves responses of the network on older tasks by using a distillation loss~\cite{hinton2015distilling}, where response targets are computed using data from the current task. As a result, LwF does not require the storage of older training data, however, this very strategy can cause issues if the data for the new task belongs to a distribution different from that of prior tasks. As more dissimilar tasks are added to the network, the performance on the prior tasks degrades rapidly~\cite{li2016learning}. EWC tries to minimize the change in weights that are important to previous tasks through the use of a quadratic constraint that tries to ensure that they do not stray too far from their initial values. 
Similar to LwF and EWC, we do not require the storage of older data. Like EWC, we want to avoid changing weights that are important to the prior tasks. We, however, do not use a soft constraint, but employ network pruning techniques 
to identify the most important parameters, as explained shortly.
In contrast to these prior works, adding even a very unrelated new task using our method does not change performance on older tasks at all.

As neural networks have become deeper and larger, a number of works have emerged aiming to reduce the size of trained models, as well as the computation required  for inference, either by reducing the numerical precision required for storing the network weights~\cite{gupta2015deep,han2015deep,hubara2016binarized,rastegari2016xnor}, or by pruning unimportant network weights~\cite{han2016dsd,han2015learning,li2016pruning,liu2017learning,molchanov2016pruning}. 
Our key idea is to use network pruning methods to free up parameters in the network, and then use these parameters to learn a new task. We adopt the simple weight-magnitude-based pruning method introduced in~\cite{han2016dsd,han2015learning} as it is able to prune over 50\% of the parameters of the initial network. As we will discuss in Section~\ref{subsec:results_filter_pruning}, we also experimented with the filter-based pruning of~\cite{molchanov2016pruning}, obtaining limited success due to the inability to prune aggressively. 
Our work is related to the very recent method proposed by Han \etal~\cite{han2016dsd}, which shows that sparsifying and retraining weights of a network serves as a form of regularization and improves performance {\em on the same task}. In contrast, we use iterative pruning and re-training to add multiple diverse tasks. 


It is possible to limit performance loss on older tasks if one allows the network to grow as new tasks are added. One approach, called progressive neural networks~\cite{rusu2016progressive}, replicates the network architecture for every new dataset, with each new layer augmented with lateral connections to corresponding older layers. The weights of the new layers are optimized, while keeping the weights of the old layers frozen. The initial networks are thus unchanged, while the new layers are able to re-use representations from the older tasks. One unavoidable drawback of this approach is that the size of the full network keeps increasing with the number of added tasks. The overhead per dataset added for our method is lower than in~\cite{rusu2016progressive} as we only store one binary parameter selection mask per task, which can further be combined across tasks, as explained in the next section.
Another recent idea, called PathNet~\cite{fernando2017pathnet}, 
uses evolutionary strategies to select pathways through the network. They too, freeze older pathways while allowing newly introduced tasks to re-use older neurons. At a high hevel, our method aims at achieving similar behavior, but without resorting to computationally intensive search over architectures or pathways.

To our knowledge, our work presents the most extensive set of experiments on full-scale real image datasets and state-of-the-art architectures to date.
Most existing work on transfer and multi-task learning, like~\cite{fernando2017pathnet,kirkpatrick2017overcoming,lee2017overcoming,rusu2016progressive}, performed validation on small-image datasets (MNIST, CIFAR-10) or synthetic reinforcement learning environments (Atari, 3D maze games). Experiments with EWC and LwF have demonstrated the addition of just one task, or subsets of the same dataset~\cite{lee2017overcoming,li2016learning}.
By contrast, we demonstrate the successful combination of up to four tasks in a single network: starting with an ImageNet-trained VGG-16 network, we sequentially add three fine-grained classification tasks on CUBS birds~\cite{WahCUB_200_2011}, Stanford Cars~\cite{krause20133d}, and Oxford Flowers~\cite{Nilsback08} datasets. We also combine ImageNet classification with scene classification on the Places365~\cite{zhou2017places} dataset that has 1.8M training examples. In all experiments, our method achieves performance close to the best possible case of using one separate network per task. Further, we show that our pruning-based scheme generalizes to architectures with batch normalization~\cite{ioffe2015batch}, residual connections~\cite{he2016deep}, and dense connections~\cite{huang2017densely}.

Finally, our work is related to incremental learning approaches~\cite{rebuffi2016icarl,shmelkov2017incremental}, which focus on the addition of classifiers or detectors for a few classes at a time. Our setting differs from theirs in that we explore the addition of entire image classification tasks or entire datasets at once.



\section{Approach}
\label{sec:approach}

The basic idea of our approach is to use network pruning techniques to create free parameters that can then be employed for learning new tasks, without adding extra network capacity. 


\smallskip
\noindent{\bf Training.} Figure~\ref{fig:filters_illustration} gives an overview of our method. We begin with a standard network learned for an initial task, such as the VGG-16~\cite{simonyan14VGG} trained on  ImageNet~\cite{ILSVRC15} classification, referred to as Task I. The initial weights of a filter are depicted in gray in Figure~\ref{fig:filters_illustration} (a).
We then prune away a certain fraction of the weights of the network, \ie set them to zero.
Pruning a network results in a loss in performance due to the sudden change in network connectivity. This is especially pronounced when the pruning ratio is high.
In order to regain accuracy after pruning, we need to re-train the network for a smaller number of epochs than those required for training.
After a round of pruning and re-training, we obtain a network with sparse filters and minimal reduction in performance on Task I. 
The surviving parameters of Task I, those in gray in Figure~\ref{fig:filters_illustration} (b), are hereafter kept fixed.

Next, we train the network for a new task, Task II, and let the pruned weights come back from zero, obtaining orange colored weights as shown in Figure~\ref{fig:filters_illustration} (c). Note that the filter for Task II makes use of both the gray and orange weights, \ie weights belonging to the previous task(s) are re-used.
We once again prune the network, freeing up some parameters used for Task II only, and re-train for Task II to recover from pruning. 
This gives us the filter illustrated in Figure~\ref{fig:filters_illustration} (d).
At this point onwards, the weights for Tasks I and II are kept fixed.
The available pruned parameters are then employed for learning yet another new task, resulting in green-colored weights shown in Figure~\ref{fig:filters_illustration} (e). This process is repeated until all the required tasks are added or no more free parameters are available. In our experiments, pruning and re-training is about $1.5\times$ longer than simple fine-tuning, as we generally re-train for half the training epochs.

\smallskip
\noindent{\bf Pruning Procedure.}
In each round of pruning, we remove a fixed percentage of eligible weights from every convolutional and fully connected layer. The weights in a layer are sorted by their absolute magnitude, and the lowest 50\% or 75\% are selected for removal, similar to~\cite{han2016dsd}. We use a one-shot pruning approach for simplicity, though incremental pruning has been shown to achieve better performance~\cite{han2015learning}.
As previously stated, we only prune weights belonging to the current task, and do not modify weights that belong to a prior task. For example, in going from filter (c) to (d) in Figure~\ref{fig:filters_illustration}, we only prune from the orange weights belonging to Task II, while gray weights of Task I remain fixed. This ensures no change in performance on prior tasks while adding a new task. 


We did not find it necessary to learn task-specific biases similar to EWC~\cite{kirkpatrick2017overcoming}, and keep the biases of all the layers fixed after the network is pruned and re-trained for the first time. Similarly, in networks that use batch normalization, we do not update the parameters (gain, bias) or running averages (mean, variance), after the first round of pruning and re-training. This choice  helps reduce the additional per-task overhead, and it is justified by our results in the next section and further analysis performed in Section~\ref{sec:analysis}.

The only overhead of adding multiple tasks is the storage of a sparsity mask indicating which parameters are active for a particular task. 
By following the iterative training procedure, for a particular Task $K$, we obtain a filter that is  the superposition of weights learned for that particular task and weights learned for all previous Tasks $1,\cdots, K-1$. 
If a parameter is first used by Task $K$, it is used by all tasks $K,\cdots,N$, where $N$ is the total number of tasks.
Thus, we need at most $\log_2(N)$ bits to encode the mask per parameter, instead of 1 bit per task, per parameter. The overhead for adding one and three tasks to the initial ImageNet-trained VGG-16 network (\texttt{conv1\_1} to \texttt{fc\_7}) of size 537 MB is only $\sim$17 MB and $\sim$34 MB, respectively. A network with four tasks total thus results in a 1/16 increase with respect to the initial size, as a typical parameter is represented using 4 bytes, or 32 bits.\footnote{In practice, we store masks inside a PyTorch ByteTensor (1 byte = 8 bits) due to lack of support for arbitrary-precision storage.}

\smallskip
\noindent{\bf Inference.}
When performing inference for a selected task, the network parameters are masked so that the network state matches the one learned during training, \ie the filter from Figure~\ref{fig:filters_illustration} (b) for inference on Task I, Figure~\ref{fig:filters_illustration} (d) for inference on Task II, and so on.
There is no additional run-time overhead as no extra computation is required; weights only have to be masked in a binary on/off fashion during multiplication, which can easily be implemented in the matrix-matrix multiplication kernels.

It is important to note that our pruning-based method is unable to perform simultaneous inference on all tasks as responses of a filter change depending on its level of sparsity, and are no longer separable after passing through a non-linearity such as the ReLU. 
Performing filter-level pruning, in which an entire filter is switched on/off, instead of a single parameter, can allow for simultaneous inference. However, we show in Section~\ref{subsec:results_filter_pruning} that such methods are currently limited in their pruning ability and cannot accommodate multiple tasks without significant loss in performance.


\section{Experiments and Results}
\label{sec:results}

\noindent{\bf Datasets and Training Settings.}
We evaluate our method on two large-scale image datasets and three fine-grained classification datasets, as summarized in Table~\ref{table:dataset_stats}. 

\begin{table}[th!]
  \centering
  \begin{tabular}{|l|c|c|c|}
    \hline
    {\bf Dataset} & {\bf \#Train} & {\bf \#Eval} & {\bf \#Classes} \\\hline\hline
    ImageNet~\cite{ILSVRC15} & 1,281,144 & 50,000 & 1,000 \\\hline
    Places365~\cite{zhou2017places} & 1,803,460 & 36,500 & 365 \\\hline
    CUBS Birds~\cite{WahCUB_200_2011} & 5,994 & 5,794 & 200 \\\hline
    Stanford Cars~\cite{krause20133d} & 8,144 & 8,041 & 196 \\\hline
    Flowers~\cite{Nilsback08} & 2,040 & 6,149 & 102 \\\hline
  \end{tabular}
  \caption{Summary of datasets used. }
  \label{table:dataset_stats}
\end{table}

In the case of the Stanford Cars and CUBS datasets, we crop object bounding boxes out of the input images and resize them to $224\times 224$. For the other datasets, we resize the input image to $256\times 256$  and take a random crop of size $224\times 224$ as input. For all datasets, we perform left-right flips for data augmentation.

In all experiments, we begin with an ImageNet-trained network, as it is essential to have a good starting set of parameters. The only change we make to the network is the addition of a new output layer per each new task.
After pruning the initial ImageNet-trained network, we fine-tune it on the ImageNet dataset for 10 epochs with a learning rate of 1e-3 decayed by a factor of 10 after 5 epochs. For adding fine-grained datasets, we use the same initial learning rate, decayed after 10 epochs, and train for a total of 20 epochs. For the larger Places365 dataset, we fine-tune for a total of 10 epochs, with learning rate decay after 5 epochs. 
When a network is pruned after training for a new task, we further fine-tune the network for 10 epochs with a constant learning rate of 1e-4. 
We use a batch size of 32 and the default dropout rates on all networks.

\begin{table*}[ht!]
  \centering
  \begin{tabular}{|l||c||c||c|c||c|}
    \hline
    \multirow{2}{*}{\bf Dataset} & {\bf Classifier} & \multirow{2}{*}{\bf LwF} & \multicolumn{2}{c||}{\bf Pruning (ours)} & {\bf Individual} \\ 
    & {\bf Only} & & 0.50, 0.75, 0.75 & 0.75, 0.75, 0.75 & {\bf Networks} \\\hline\hline
    \multirow{2}{*}{ImageNet} & 28.42 & 39.23 & 29.33 & 30.87 & 28.42 \\
    & (9.61) & (16.94) & (9.99) & (10.93) & (9.61) \\\hline
    CUBS & 36.76 & 30.42 & 25.72 & 24.95 & 22.57 \\\hline
    Stanford Cars & 56.42 & 22.97 & 18.08 & 15.75 & 13.97 \\\hline
    Flowers & 20.50 & 15.21 & 10.09 & 9.75 & 8.65 \\\hline\hline
    \# Models (Size) & 1 (562 MB) & 1 (562 MB) & 1 (595 MB) & 1 (595 MB) & {4 (2,173 MB)} \\\hline
  \end{tabular}
  \caption{Errors on fine-grained tasks. Values in parentheses are top-5 errors, while all others are top-1 errors. The numbers at the top of the Pruning columns indicate the ratios by which the network is pruned after each successive task. For example, 0.50, 0.75, 0.75 indicates that the initial ImageNet-trained network is pruned by 50\%, and after each task is added, 75\% of the parameters belonging to that task are set to 0. The results in the Pruning columns are averaged over 18 runs with varying order of training of the 3 datasets (6 possible orderings, 3 runs per ordering), and those in the LwF column are over 1 run per ordering. Classifier Only and Individual Network values are averaged over 3 runs. }
  \vspace{10pt}
  \label{table:results}
\end{table*}

\begin{table*}[th!]
  \centering
  \begin{tabular}{|l||c||c|c||c|}
    \hline
    \multirow{2}{*}{\bf Dataset} & {\bf Jointly Trained} & \multicolumn{2}{c||}{\bf Pruning (ours)} & {\bf Individual} \\ 
    & {\bf Network$^\ast$} & 0.50 & 0.75 & {\bf Networks} \\\hline\hline
    \multirow{2}{*}{ImageNet} & 33.49 & 29.33 & 30.87 & 28.42 \\
    & (12.25) & (9.99) & (10.93) & (9.61) \\\hline
    \multirow{2}{*}{Places365} & 45.98 & 47.44 & 46.99 & 46.35 \\
    & (15.59) & (16.67) & (16.24) & (16.14)$^\ast$ \\\hline\hline
    \# Models (Size) & 1 (559 MB) & 1 (576 MB) & 1 (576 MB) & {2 (1,096 MB)} \\\hline
  \end{tabular}
  \caption{Results when an ImageNet-trained VGG-16 network is pruned by 50\% and 75\% and the Places dataset is added to it. Values in parentheses are top-5 errors, while all others are top-1 errors. $^\ast$ indicates models downloaded from \url{https://github.com/CSAILVision/places365}, trained by~\cite{zhou2017places}.}
  \label{table:results_places}
\end{table*}

\smallskip
\noindent{\bf Baselines.}
The simplest baseline method, referred to as {\bf Classifier Only}, is to extract the \texttt{fc7} or pre-classifier features from the initial network and only train a new classifier for each specific task, meaning that the performance on ImageNet remains the same. For training each new classifier layer, we use a constant learning rate of 1e-3 for 20 epochs. 

The second baseline, referred to as {\bf Individual Networks}, 
trains separate models for every task, achieving the highest possible accuracies by dedicating all the resources of the network for that single task. To obtain models for individual fine-grained tasks, we start with the ImageNet-trained network and fine-tune on the respective task for 20 epochs total with a learning rate of 1e-3 decayed by factor of 10 after 10 epochs.

Another baseline used in prior work~\cite{li2016learning,rannen2017encoder} is {\bf Joint Training} of a network for multiple tasks. However, joint fine-tuning is rather tricky when dataset sizes are different (\eg ImageNet and CUBS), so we do not attempt it for our experiments with fine-grained datasets, especially since individually trained networks provide higher reference accuracies in any case. Joint training works better for similarly-sized datasets, thus, when combining ImageNet and Places, we compare with the jointly trained network provided by the authors of~\cite{zhou2017places}. 

Our final baseline is our own re-implementation of {\bf LwF}~\cite{li2016learning}. We use the same default settings as in~\cite{li2016learning}, including a unit tradeoff parameter between the distillation loss and the loss on the training data for the new task. For adding fine-grained datasets with LwF, we use an initial learning rate of 1e-3 decayed after 10 epochs, and train for a total of 20 epochs. In the first 5 epochs, we train only the new classifier layer, as recommended in~\cite{li2016learning}.

\smallskip
\noindent{\bf Multiple fine-grained classification tasks.} Table~\ref{table:results} summarizes the experiments in which we add the three fine-grained tasks of CUBS, Cars, and Flowers classification in varying orders to the VGG-16 network. By comparing the Classifier Only and Individual Networks columns, we can clearly see that the fine-grained tasks benefit a lot by allowing the lower convolutional layers to change, with the top-1 error on cars and birds classification dropping from 56.42\% to 13.97\%, and from 36.76\% to 22.57\% respectively.

There are a total of six different orderings in which the three tasks can be added to the initial network. The Pruning columns of Table~\ref{table:results} report the averages of the top-1 errors obtained with our method across these six orderings, with three independent runs per ordering. Detailed exploration of the effect of ordering will be presented in the next section. 
By pruning and re-training the ImageNet-trained VGG-16 network by 50\% and 75\%, the top-1 error slightly increases from the initial 28.42\% to 29.33\% and 30.87\%, respectively, and the top-5 error slightly increases from 9.61\% to 9.99\% and 10.93\%. 
When three tasks are added to the 75\% pruned initial network, we achieve errors
CUBS, Stanford Cars, and Flowers that are only 2.38\%, 1.78\%, and 1.10\% worse than the Individual Networks best case. At the same time, the errors are reduced by 11.04\%, 30.41\%, and 10.41\% compared to the Classifier Only baseline.
Not surprisingly, starting with a network that is initially pruned by a higher ratio results in better performance on the fine-grained tasks, as it makes more parameters available for them. This especially helps the challenging Cars classification, reducing top-1 error from 18.08\% to 15.75\% as the initial pruning ratio is increased from 50\% to 75\%. 

\begin{figure*}[t!]
  \centering
  \subfigure{\includegraphics[trim={0.4cm 0.8cm 0.3cm 0cm}, width=\columnwidth]{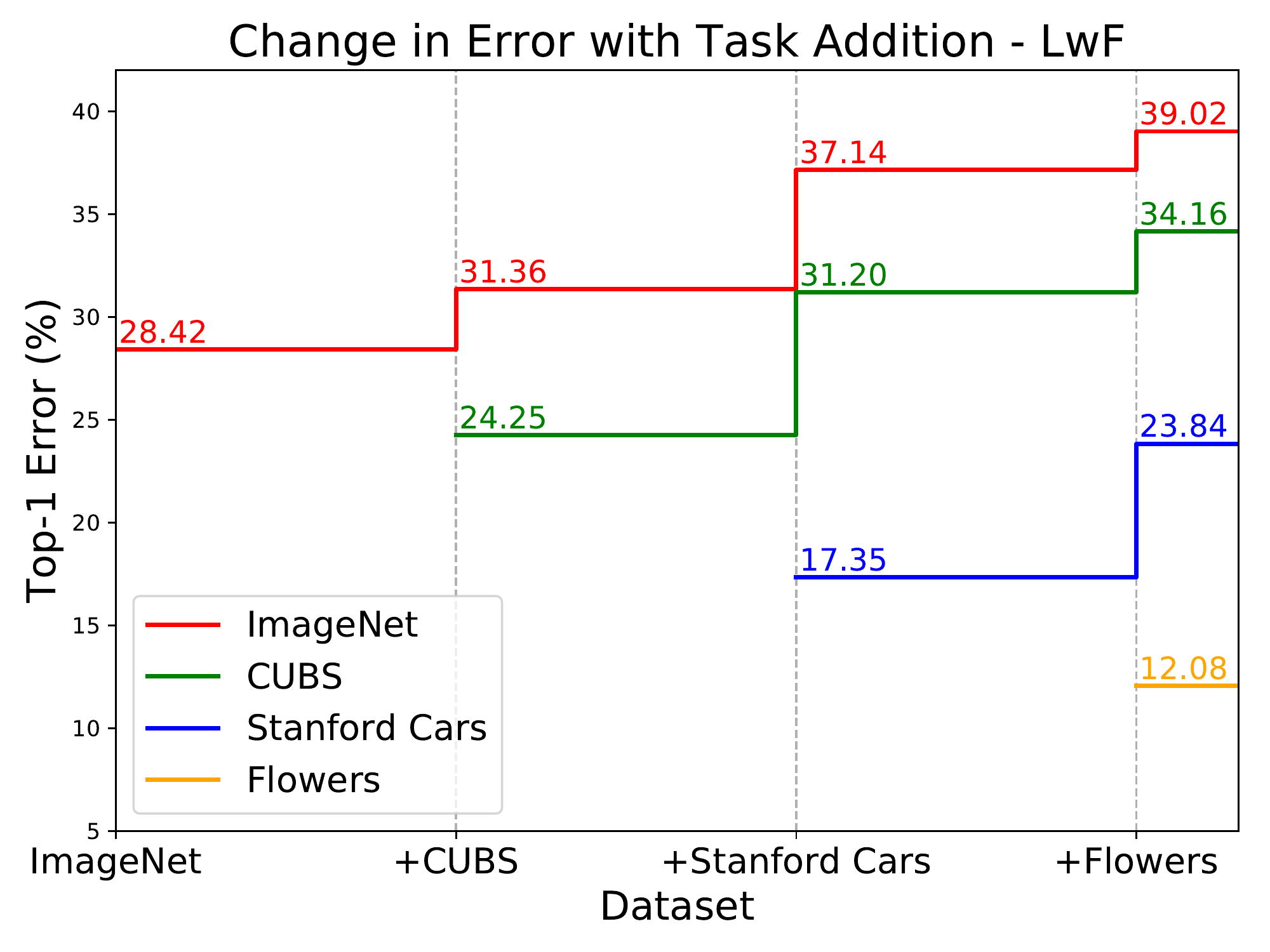}}
  \hfill
  \subfigure{\includegraphics[trim={0.4cm 0.8cm 0.3cm 0cm}, width=\columnwidth]{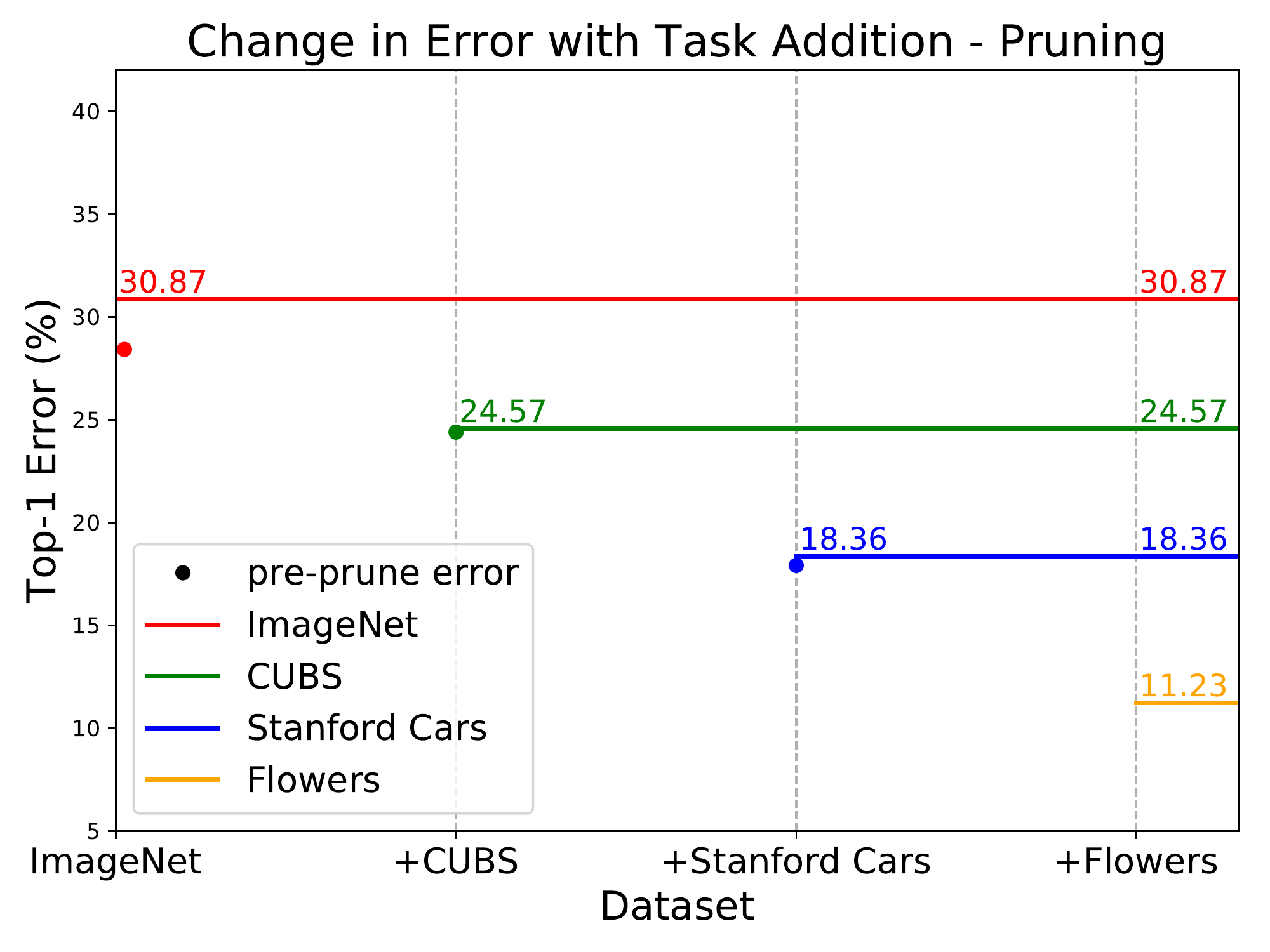}}
  \caption{Change in errors on prior tasks as new tasks are added for LwF (left) and our method (right). For LwF, errors on prior datasets increase with every added dataset. For our pruning-based method, the error remains the same even after a new dataset is added.}
  \label{fig:change_with_addition}
\end{figure*}

Our approach also consistently beats LwF on all datasets. 
As seen in Figure~\ref{fig:change_with_addition}, while training for a new task, the error on older tasks increases continuously in the case of LwF, whereas it remains fixed for our method. The unpredictable change in older task accuracies for LwF is problematic, especially when we want to guarantee a specific level of performance.
 
Finally, as shown in the last row of Table~\ref{table:results}, our pruning-based model is much smaller than training separate networks per task (595 MB v/s 2,173 MB), and is only 33 MB larger than the classifier-only baseline.

\smallskip
\noindent{\bf Adding another large-scale dataset task.} Table~\ref{table:results_places} shows the results of adding the large-scale Places365 classification task to a pruned ImageNet network. By adding Places365, which is larger than ImageNet (1.8 M images v/s 1.3 M images), to a 75\% pruned ImageNet-trained network, we achieve top-1 error within 0.64\% and top-5 error within 0.10\% of an individually trained network. By contrast, the jointly trained baseline obtains performance much worse than an individual network for ImageNet (33.49\% v/s 28.42\% top-1 error). This highlights a common problem associated with joint training, namely, the need to balance mixing ratios between the multiple datasets which may or may not be complementary, and accommodate their possibly different hyperparamter requirements. In comparison, iterative pruning allows for a controlled decrease in prior task performance and for the use of different training hyperparameter settings per task. Further, we trained the pruned network on Places365 for 10 epochs only, while the joint and individual networks were trained for 60-90 epochs~\cite{zhou2017places}.

\begin{table}
  \resizebox{\columnwidth}{!}{%
  \begin{tabular}{|l||c||c||c|}
    \hline
    \multirow{2}{*}{\bf Dataset} & {\bf Classifier} & {\bf Pruning (ours)} & {\bf Individual} \\
    & {\bf Only} & {\bf 0.50, 0.75, 0.75} & {\bf Networks} \\\hline
    \multicolumn{4}{|c|}{\bf VGG-16 with Batch Normalization} \\\hline
    \multirow{2}{*}{ImageNet} & 26.63 & 27.18 & 26.63 \\
    & (8.49) & (8.69) & (8.49) \\\hline
    CUBS & 35.26 & 21.89 & 19.83 \\\hline
    Stanford Cars & 57.21 & 14.57 & 13.29 \\\hline
    Flowers & 21.79 & 7.45 & 6.04 \\\hline\hline
    Size & 562 MB & 595 MB & 2,173 MB \\\hline\hline
    \multicolumn{4}{|c|}{\bf ResNet-50} \\\hline
    \multirow{2}{*}{ImageNet} & 23.84 & 24.29 & 23.84 \\
    & (7.13) & (7.18) & (7.13) \\\hline
    CUBS & 34.83 & 21.13 & 19.56 \\\hline
    Stanford Cars & 58.15 & 13.75 & 12.99 \\\hline
    Flowers & 18.53 & 7.10 & 8.50 \\\hline\hline
    Size & 107 MB & 112 MB & 389 MB \\\hline\hline
    \multicolumn{4}{|c|}{\bf DenseNet-121} \\\hline
    \multirow{2}{*}{ImageNet} & 25.56 & 25.60 & 25.56 \\
    & (8.02) & (7.89) & (8.02) \\\hline
    CUBS & 28.88 & 21.84 & 19.72 \\\hline
    Stanford Cars & 47.65 & 15.55 & 13.15 \\\hline
    Flowers & 17.12 & 7.71 & 8.02 \\\hline\hline
    Size & 34 MB & 36 MB & 119 MB \\\hline
  \end{tabular}
  }
  \caption{Results on additional network types. Values in parentheses are top-5 errors, while all others are top-1 errors. The results in the pruning column are averaged over 18 runs with varying order of training of the 3 datasets (6 possible orderings, 3 runs per ordering). Classifier Only and Individual Network values are averaged over 3 runs.}
  \label{table:other_results}
\end{table}

\smallskip
\noindent{\bf Extension to other networks.} The results presented so far were obtained for the vanilla VGG-16 network, a simple and large network, well known to be full of redundancies~\cite{canziani2016analysis}. Newer architectures such as ResNets~\cite{he2016deep} and DenseNets~\cite{huang2017densely} are much more compact, deeper, and better-performing. For comparison, the Classifier Only models of VGG-16, ResNet-50, and DenseNet-121 have 140 M, 27 M, and 8.6 M parameters respectively. It is not obvious how well pruning will work on the latter two parameter-efficient networks. Further, one might wonder whether sharing batch normalization parameters across diverse tasks might limit accuracy.
Table~\ref{table:other_results} shows that our method can indeed be applied to all these architectures, which include residual connections, skip connections, and batch normalization.
As described in Section~\ref{sec:approach}, the batch normalization parameters (gain, bias, running means, and variances) are frozen after the network is pruned and retrained for ImageNet. In spite of this constraint, we achieve errors much lower than the baseline that only trains the last classifier layer. In almost all cases, we obtain errors within 1-2\% of the best case scenario of one network per task. While we tried learning separate batchnorm parameters per task and this further improved performance, we chose to freeze batchnorm parameters since it is simpler and avoids the overhead of storing these separate parameters (4 vectors per batchnorm layer).

The deeper ResNet and DenseNet networks with 50 and 121 layers, respectively, are very robust to pruning, losing just 0.45\% and 0.04\% top-1 accuracy on ImageNet, respectively.  Top-5 error increases by 0.05\% for ResNet, and decreases by 0.13\% for DenseNet. 
In the case of Flowers classification, we perform better than the individual network, probably because training the full network causes it to overfit to the Flowers dataset, which is the smallest. By using the fewer available parameters after pruning, we likely avoid this issue.


Apart from obtaining good performance across a range of networks, an additional benefit of our pruning-based approach is that for a given task, the network can be pruned by small amounts iteratively so that the desirable trade-off between loss of current task accuracy and provisioning of free parameters for subsequent tasks can be achieved. Note that the fewer the parameters, the lower the mask storage overhead of our methods, as seen in the Size rows of Table~\ref{table:other_results}.


\section{Detailed Analysis}
\label{sec:analysis}

In this section, we investigate the factors that affect performance while using our method, 
and justify choices made such as freezing biases of the network. 
We also compare our weight-pruning approach with a filter-pruning approach, and confirm its benefits over the latter.

\subsection{Effect of training order}
\label{subsec:results_order}

\begin{figure}[t!]
\includegraphics[trim={0.4cm 0.4cm 0.4cm 0cm}, width=\columnwidth]{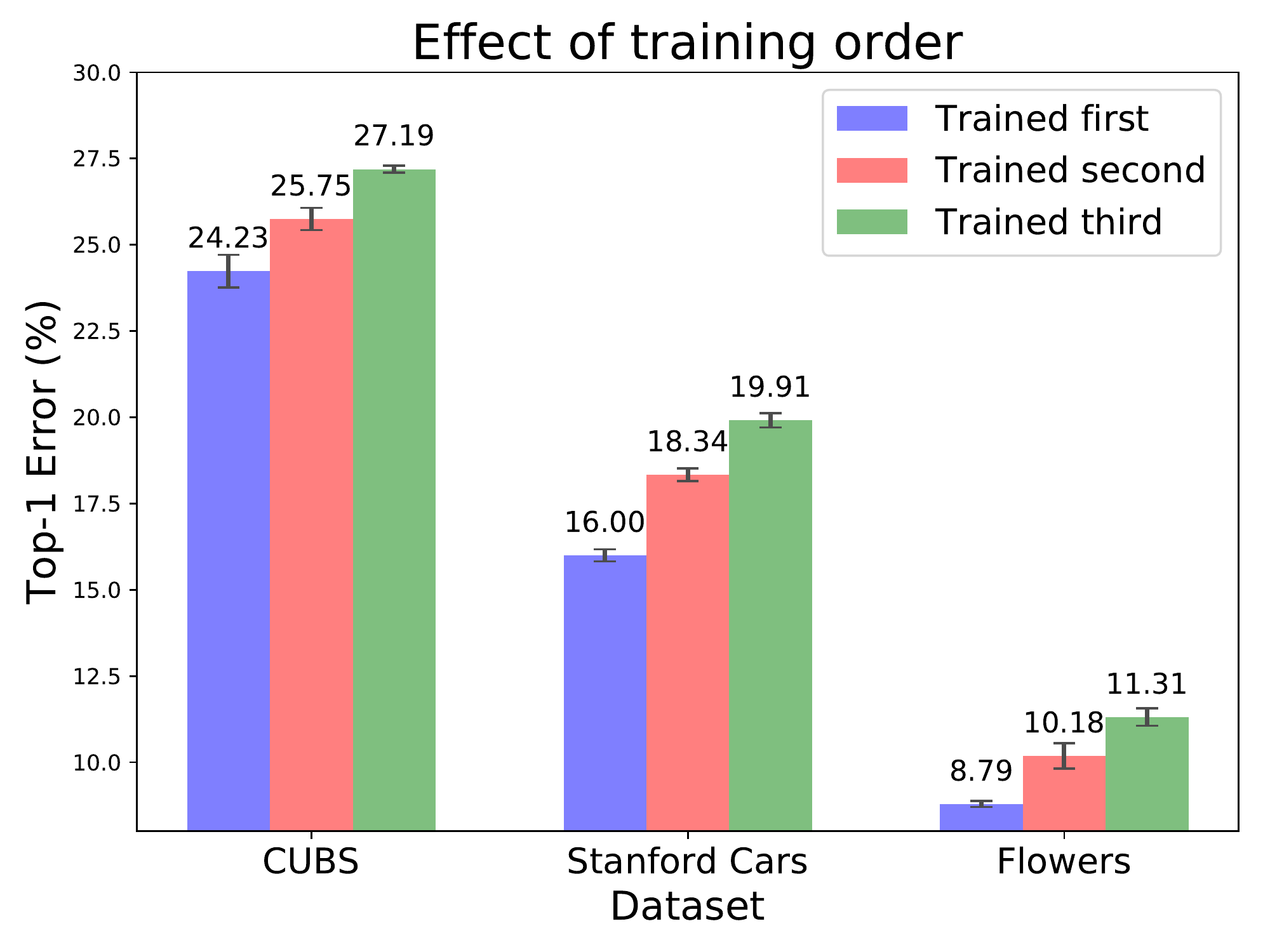}
\caption{Dependence of errors on individual tasks on the order of task addition (see text for details). Each displayed value and error bar are obtained from 6 different runs. We use an initial pruning ratio of 50\% for the ImageNet-trained VGG-16 and a pruning ratio of 75\% after each dataset is added. 0.50, 0.75, 0.75 pruning column of Table~\ref{table:results} reports the average over orderings.}
\label{fig:training_order_nobias}
\end{figure}

As more tasks are added to a network, a larger fraction of the network becomes unavailable for tasks that are subsequently added. Consider the 0.50, 0.75, 0.75 pruning ratio sequence for the VGG-16 network. The layers from \texttt{conv1\_1} to \texttt{fc\_7} contain around 134 M parameters. After the initial round of 50\% pruning for Task I (ImageNet classification), we have $\sim$67 M free parameters. After the second round of training followed by 75\% pruning and re-training, 16.75 M parameters are used by Task II, and 50.25 M free parameters available for subsequent tasks. Likewise, Task III uses around 13 M parameters and leaves around 37 M free parameters for Task IV. Accordingly, we observe a reduction of accuracy with order of training, as shown in Figure~\ref{fig:training_order_nobias}. For example, the top-1 error increases from 16.00\% to 18.34\% to 19.91\% for the Stanford Cars dataset as we delay its addition to the network. For the datasets considered, the error increases by 3\% on average when the order of addition is changed from first to third.
Note that the results reported in Table~\ref{table:results} are averaged over all orderings for a particular dataset.
These findings suggest that if it is possible to decide the ordering of tasks beforehand, the most challenging or unrelated task should be added first.

\subsection{Effect of pruning ratios}
\label{subsec:results_biases}

\begin{figure}[t!]
\includegraphics[trim={0.4cm 0.4cm 0.4cm 0cm}, width=\columnwidth]{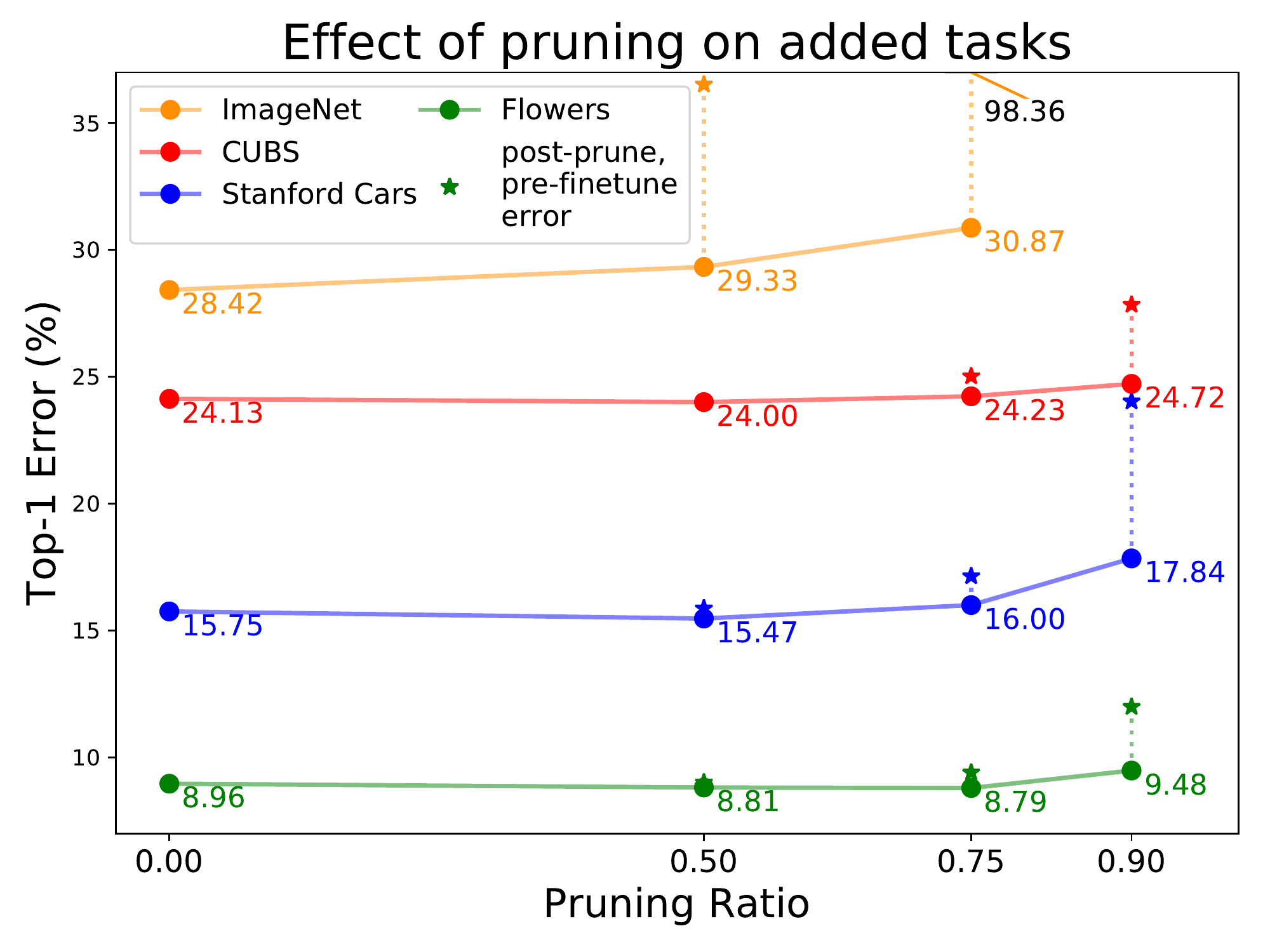}
\caption{This plot measures the change in top-1 error with pruning. The values above correspond to the case when the respective dataset is added as the first task, to an ImageNet-trained VGG-16 that is 50\% pruned, except for the values corresponding to the ImageNet dataset which correspond to initial pruning. Note that the 0.75 pruning ratio values correspond to the blue bars in Figure~\ref{fig:training_order_nobias}.}
\label{fig:drop_after_pruning}
\end{figure}

In Figure~\ref{fig:drop_after_pruning}, we measure the effect of pruning and re-training for a task, when it is first added to a 50\% pruned VGG-16 network (except for the initial ImageNet task). We consider this specific case in order to isolate the effect of pruning from the order of training discussed above. We observe that the errors for a task increase immediately upon pruning ($\star$ markers) due to sudden change in network connectivity. However, upon re-training, the errors reduce, and might even drop below the original unpruned error, as seen for all datasets other than ImageNet at the 50\% pruning ratio, in line with prior work~\cite{han2016dsd} which has shown that pruning and retraining can function as effective regularization. Multi-step pruning will definitely help reduce errors on ImageNet, as reported in~\cite{han2015learning}.
This plot shows that re-training is essential, especially when the pruning ratios are large. 

Interestingly, for a newly added task, 50\% and 75\% pruning without re-training does not increase the error by much.
More surprisingly, even a very aggressive single-shot pruning ratio of 90\% followed by re-training results in a small error increase compared to the unpruned errors (top-1 error increases from 15.75\% to 17.84\% for Stanford Cars, 24.13\% to 24.72\% for CUBS, and 8.96\% to 9.48\% for Flowers). This indicates effective transfer learning as very few parameter modifications (10\% of the available 50\% of total parameters after pruning, or 5\% of the total VGG-16 parameters) are enough to obtain good accuracies.

\subsection{Effect of training separate biases}
\label{subsec:results_biases}
We do not observe any noticeable improvement in performance by learning task-specific biases per layer, as shown in Table~\ref{table:bias_comparison}. Sharing biases reduces the storage overhead of our proposed method, as each convolutional, fully-connected, or batch-normalization layer can contain an associated bias term. We thus choose not to learn task-specific biases in our reported results.
\begin{table}[h!]
  \centering
  \begin{tabular}{|l||C{2cm}|C{2cm}|}
    \hline
    \multirow{2}{*}{\bf Dataset} & \multicolumn{2}{c|}{\bf Pruning 0.50, 0.75, 0.75} \\
     & Separate Bias & Shared Bias  \\\hline\hline
     CUBS & 25.62 & 25.72 \\\hline
     Stanford Cars & 18.17 & 18.08 \\\hline
     Flowers & 10.11 & 10.09  \\\hline
  \end{tabular}
  \caption{No noticeable difference in performance is observed by learning task-specific biases. Values are averaged across all 6 task orderings, with 3 runs per ordering. The shared bias column corresponds to the 0.50, 0.75, 0.75 Pruning column of Table~\ref{table:results}.}
  \label{table:bias_comparison}
\end{table}


\subsection{Is training of all layers required?}
\label{subsec:results_layers}
\begin{figure}[t!]
\includegraphics[trim={0.4cm 0.4cm 0.4cm 0cm}, width=\columnwidth]{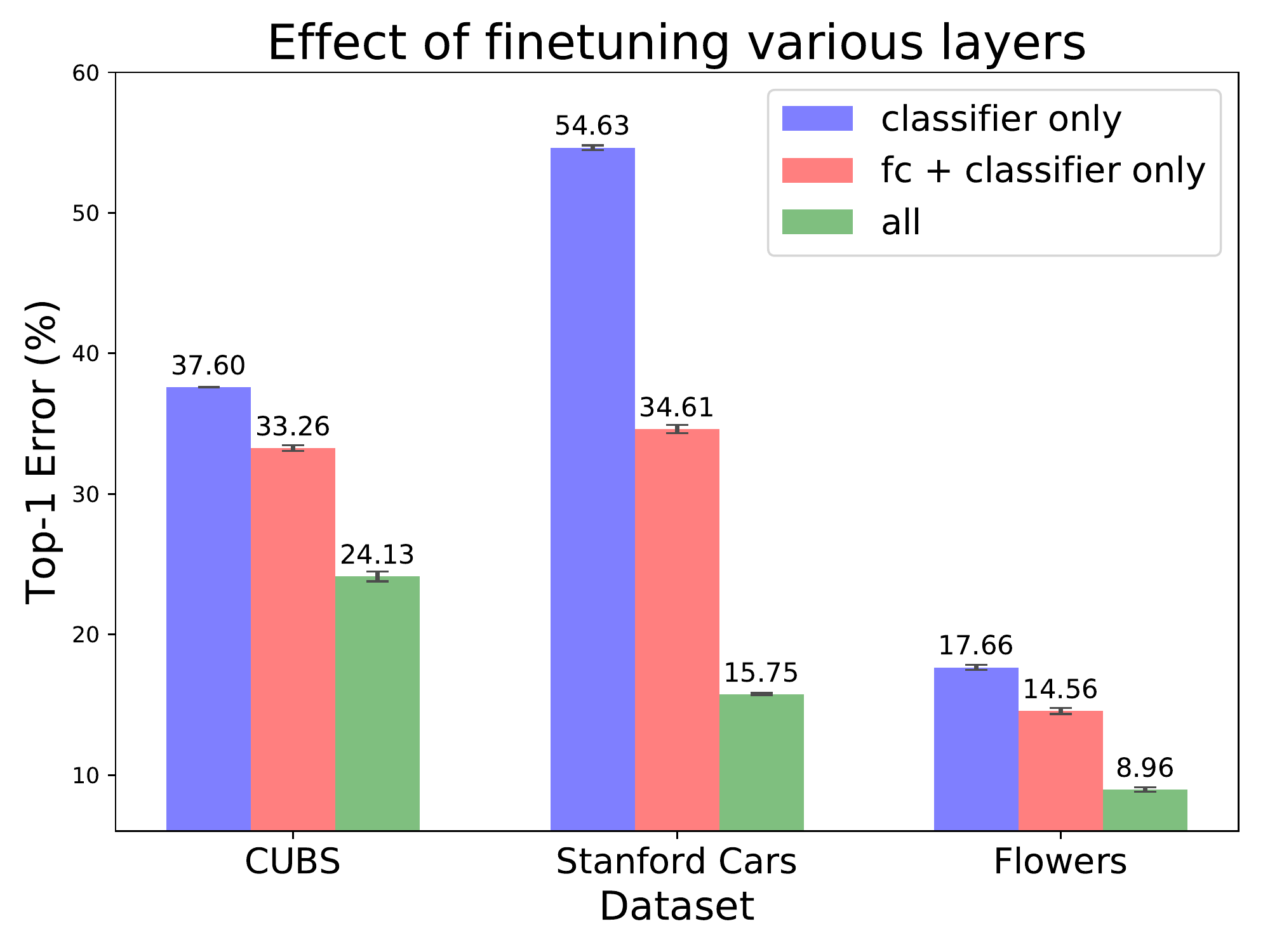}
\caption{This figure shows that having free parameters in the lower layers of the network is essential for good performance. The numbers above are obtained when a task is added to the 50\% pruned VGG-16 network and the only the specified layers are finetuned, without any further pruning.}
\label{fig:training_layers_ft}
\end{figure}

Figure~\ref{fig:training_layers_ft} measures the effect of modifying freed-up parameters from various layers for learning a new task. For this experiment, we start with the 50\% pruned ImageNet-trained vanilla VGG-16 network, and add one new task. For the new task, we train pruned neurons from the specified layers only. Fine-tuning the fully connected layers improves accuracy over the classifier only baseline in all tasks. Further, fine-tuning the convolutional layers provides the biggest boost in accuracy, and is clearly necessary for obtaining good performance. By using our method, we can control the number of pruned parameters at each layer, allowing one to make use of task-specific requirements, when available.

\subsection{Comparison with filter-based pruning}
\label{subsec:results_filter_pruning}
For completeness, we report experiments with filter-based pruning~\cite{molchanov2016pruning}, which eliminates entire filters, instead of sparsifying them. The biggest advantage of this strategy is that it enables simultaneous inference to be performed for all the trained tasks.
For filters that survive a round of pruning, incoming weights on all filters that did not survive pruning (and are hence available for subsequent tasks) are set to 0. As a result, when new filters are learned for a new task, their outputs would not be used by filters of prior tasks. Thus, the output of a filter for a prior task would always remain the same, irrespective of filters learned for tasks added later.
The method of~\cite{molchanov2016pruning} ranks all filters in a network based on their importance to the current dataset, as measured by a metric related to the Taylor expansion of the loss function. We prune 400 filters per each epoch of $\sim$40,000 iterations, for a total of 10 epochs. 
Altogether, this eliminates 4,000 filters from a total of 12,416 in VGG-16, or $\sim$30\% pruning. We could not prune more aggressively without substantially reducing accuracy on ImageNet. A further unfavorable observation is that most of the pruned filters (3,730 out of 4,000) were chosen from the fully connected layers (Liu \etal~\cite{liu2017learning} proposed a different filter-based pruning method and found similar behavior for VGG-16). This frees up too few parameters in the lower layers of the network to be able to fine-tune effectively for new tasks. As a result, filter-based pruning only allowed us to add one extra task to the ImageNet-trained VGG-16 network, as shown in Table~\ref{table:filter_pruning_comparison}. 
\begin{table}[t!]
  \centering
  \begin{tabular}{|l||c||c|c|}
    \hline
    \multirow{2}{*}{\bf Dataset} & {\bf Classifier} & \multicolumn{2}{c|}{\bf Pruning} \\
    & {\bf Only} & Filters & Weights \\
    \hline\hline
    \multirow{2}{*}{ImageNet} & 28.42 & 30.70 & {\bf 29.33} \\
    & (9.61) & (10.92) & ({\bf 9.99}) \\\hline
    CUBS & 36.76 & 35.73 & {\bf 24.23}  \\\hline
    Stanford Cars & 56.42 & 34.78 & {\bf 13.97}  \\\hline
    Flowers & 20.50 & 13.31 & {\bf 8.79}  \\\hline
  \end{tabular}
  \caption{Comparison of filter-based and weight-based pruning for ImageNet-trained VGG-16. This table reports errors after adding only one task to the 30\% filter-pruned and 50\% weight-pruned network. Values in the Weights column correspond to the blue bars in Figure~\ref{fig:training_order_nobias}.
  Values in parentheses are top-5 errors, and the rest are top-1 errors.}
  \label{table:filter_pruning_comparison}
\end{table}
A final disadvantage of filter-based pruning methods is that they are more complicated and require careful implementation in the case of residual networks and skip connections, as noted by Li \etal~\cite{li2016pruning}. 


\section{Conclusion}
\label{sec:conclusion}

In this work, we have presented a method to ``pack'' multiple tasks into a single network with minimal loss of performance on prior tasks. The proposed method allows us to modify all layers of a network and influence a large number of filters and features, which is necessary to obtain accuracies comparable to those of individually trained networks for each task. It works not only for the relatively ``roomy'' VGG-16 architecture, but also for more compact parameter-efficient networks such as ResNets and DenseNets.

In the future, we are interested in exploring a more general framework for multi-task learning in a single network where we jointly train both the network weights and binary sparsity masks associated with individual tasks. In our current approach, the sparsity masks per task are obtained as a result of pruning, but it might be possible to learn such masks using techniques similar to those for learning networks with binary weights~\cite{hubara2016binarized,rastegari2016xnor}.

\noindent{\bf Acknowledgments:}
We would like to thank Maxim Raginsky for a discussion that gave rise to the initial idea of this paper, and Greg Shakhnarovich for suggesting the name PackNet.
This material is based upon work supported in part by the National Science Foundation under Grants No. 1563727 and 1718221.


{\small
\bibliographystyle{ieee}
\bibliography{egbib}
}
\end{document}



\title{PackNet: Adding Multiple Tasks to a Single Network by Iterative Pruning\\Supplementary Material}



\maketitle

\section{Effect of task ordering on performance}

The 3 datasets of Stanford Cars, CUBS, and Flowers can be added in 6 orderings.
Table~\ref{table:pruning_order} shows the errors obtained by adding the datasets in the specified orders. 
The averaged values in the last column correspond to the values under the Pruning: 0.50, 0.75, 0.75 column in Table 2 of the main submission.
As analyzed in Section 4.1 of the main submission, adding a task early gives better performance.
\begin{table*}[ht!]
  \centering
  \begin{tabular}{|l||c|c|c|c|c|c||c|}
    \hline
    \multirow{2}{*}{\bf Dataset} & \multicolumn{6}{c||}{\bf Pruning: 0.50, 0.75, 0.75} & {\bf Average across}\\
    & sfc & scf & fsc & fcs & csf & cfs & {\bf orderings} \\\hline\hline
    \multirow{2}{*}{\textcolor{white}{(i) } ImageNet} & \multicolumn{6}{c||}{29.33} & 29.33\\
     & \multicolumn{6}{c||}{(9.99)} & (9.99) \\\hline
    (c) CUBS & 27.27 & 25.59 & 27.11 & 25.91 & 24.57 & 23.90 & 25.72 \\\hline
    (s) Stanford Cars & 16.01 & 15.99 & 18.31 & 19.98 & 18.36 & 19.84 & 18.08\\\hline
    (f) Flowers & 9.88 & 11.39 & 8.77 & 8.81 & 11.23 & 10.48 & 10.09 \\\hline
    
  \end{tabular}
  \caption{Errors obtained using different orderings of addition of task, when using our pruning-based method. Values in brackets are top-5 errors, all others are top-1 errors.
  Results reported for each ordering are averaged across 3 independent runs, making for a total of 18 runs.}
  \label{table:pruning_order}
\end{table*}

In Table~\ref{table:lwf_order}, are the errors obtained by adding datasets in the specified orders while using Learning without Forgetting (LwF).
\begin{table*}[h!]
  \centering
  \begin{tabular}{|l||c|c|c|c|c|c||c|}
    \hline
    \multirow{2}{*}{\bf Dataset} & \multicolumn{6}{c||}{\bf LwF} & {\bf Average across}\\
    & sfc & scf & fsc & fcs & csf & cfs & {\bf orderings} \\\hline\hline
    \multirow{2}{*}{\textcolor{white}{(i) } ImageNet} & 39.22 & 39.70 & 39.54 & 38.94 & 39.02 & 38.97 & 39.23 \\
     & (16.89) & (17.29) & (17.35) & (16.74) & (16.79) & (16.62) & (16.95) \\\hline
    (c) CUBS & 26.92 & 29.53 & 27.27 & 31.38 & 34.16 & 33.24 & 30.42 \\\hline
    (s) Stanford Cars & 28.16 & 27.68 & 22.20 & 18.28 & 23.84 & 17.67 & 22.97 \\\hline
    (f) Flowers & 13.89 & 12.05 & 19.22 & 18.43 & 12.08 & 15.60 & 15.21 \\\hline
  \end{tabular}
  \caption{Errors obtained using different orderings of addition of task, when using LwF. Values in brackets are top-5 errors, all others are top-1 errors.}
  \label{table:lwf_order}
\end{table*}

Our pruning-based approach outperforms LwF in all the cases.